# Learning Content Selection Rules for Generating Object Descriptions in Dialogue

**Pamela W. Jordan**                                            PJORDAN@PITT.EDU
*Learning Research and Development Center & Intelligent Systems Program*
*University of Pittsburgh, LRDC Rm 744*
*Pittsburgh, PA 15260*

**Marilyn A. Walker**                                    M.A.WALKER@SHEFFIELD.AC.UK
*Department of Computer Science, University of Sheffield*
*Regent Court, 211 Portobello Street*
*Sheffield S1 4DP, U.K.*

## Abstract

A fundamental requirement of any task-oriented dialogue system is the ability to generate object descriptions that refer to objects in the task domain. The subproblem of content selection for object descriptions in task-oriented dialogue has been the focus of much previous work and a large number of models have been proposed. In this paper, we use the annotated COCONUT corpus of task-oriented design dialogues to develop feature sets based on Dale and Reiter's (1995) incremental model, Brennan and Clark's (1996) conceptual pact model, and Jordan's (2000b) intentional influences model, and use these feature sets in a machine learning experiment to automatically learn a model of content selection for object descriptions. Since Dale and Reiter's model requires a representation of discourse structure, the corpus annotations are used to derive a representation based on Grosz and Sidner's (1986) theory of the intentional structure of discourse, as well as two very simple representations of discourse structure based purely on recency. We then apply the rule-induction program RIPPER to train and test the content selection component of an object description generator on a set of 393 object descriptions from the corpus. To our knowledge, this is the first reported experiment of a trainable content selection component for object description generation in dialogue. Three separate content selection models that are based on the three theoretical models, all independently achieve accuracies significantly above the MAJORITY CLASS baseline (17%) on unseen test data, with the intentional influences model (42.4%) performing significantly better than either the incremental model (30.4%) or the conceptual pact model (28.9%). But the best performing models combine all the feature sets, achieving accuracies near 60%. Surprisingly, a simple recency-based representation of discourse structure does as well as one based on intentional structure. To our knowledge, this is also the first empirical comparison of a representation of Grosz and Sidner's model of discourse structure with a simpler model for any generation task.

# 1. Introduction

A fundamental requirement of any task-oriented dialogue system is the ability to generate object descriptions that refer to objects in the task domain. For example, consider the excerpt of a task-oriented dialogue from the COCONUT corpus in Figure 1 (Di Eugenio, Jordan, Thomason, & Moore, 2000). The conversants in this dialogue are attempting to (Partial solution to problem already agreed upon in prior dialogue: 3 green chairs for 50, 1 green chair for 100 and a green table in the dining room, and a blue sofa in the living room. The numbers in brackets indicate the utterance breaks for analysis of the dialogue.)

G: [36] That leaves us with 250 dollars. [37] I have *a yellow rug for 150 dollars.* [38] Do you have any other furniture left that matches for 100 dollars?"

S: [39] No, I have no furniture left that costs $100. [40] I guess you can buy *the yellow rug for $150.*

G: [41] Okay. [42] I'll buy *the rug for 150 dollars.* [43] I have *a green chair* [44] that I can buy for 100 dollars [45] that should leave us with no money.

S: [46] That sounds good. [47] Go ahead and buy *the yellow rug* and *the green chair.*

G: [48] I'll buy *the green 100 dollar chair.* [49] Design Complete?

S: [50] Sounds good, [51] do you want *the green chair* in the dining room with *the other chairs?* [52] I put *the yellow rug* in the living room. [53] Then the design is complete.

G: [54] Sounds good. [55] Hit the design complete collaboratively construct a solution for furnishing a two room house. Each conversant starts the task with a separate set of furniture items that can be used in the solution. In the process of negotiating the solution, they generate object descriptions (shown in italics) for items of furniture.

Figure 1: Excerpt of a COCONUT dialogue illustrating variable selection of attributes for object descriptions that refer to furniture items

Each furniture type in the COCONUT task domain has four associated attributes: color, price, owner and quantity. As a first step, an object description generator must decide which of these four attributes to include in an utterance, while subsequent surface generation steps decide where in the utterance the attributes will be expressed. For example, the task domain objects under discussion in the dialogue in Figure 1 are a $150 yellow rug owned by Garrett (G) and a $100 dollar green chair owned by Steve (S). In the dialogue excerpt in Figure 1, the yellow rug is first referenced in utterance 36 as *a yellow rug for 150 dollars* and then subsequently as *the yellow rug for 150 dollars, the rug for 150 dollars, the yellow rug,* where the owner attribute is sometimes realized in a separate noun phrase within the same utterance. It could also have been described by any of the following: *the rug, my rug, my yellow rug, my $150 yellow rug, the $150 rug.* The content of these object descriptions varies depending on which attributes are included. How does the speaker decide which attributes to include?

The problem of content selection for subsequent reference has been the focus of much previous work and a large number of overlapping models have been proposed that seek to explain different aspects of referring expression content selection (Clark & Wilkes-Gibbs, 1986; Brennan & Clark, 1996; Dale & Reiter, 1995; Passonneau, 1995; Jordan, 2000b) *inter alia.* The factors that these models use include the discourse structure, the attributes and attribute values used in the previous mention, the recency of last mention, the frequency of mention, the task structure, the inferential complexity of the task, and ways of determining salient objects and the salient attributes of an object. In this paper, we use a set of factors considered important for three of these models, and empirically compare the utility of these factors as predictors in a machine learning experiment in order to first establish whether the selected factors, as we represent them, can make effective contributions to the larger task of content selection for initial as well as subsequent reference. The factor sets we utilize are:

- CONTRAST SET factors, inspired by the INCREMENTAL MODEL of Dale and Reiter (1995);

- CONCEPTUAL PACT factors, inspired by the models of Clark and colleagues (Clark & Wilkes-Gibbs, 1986; Brennan & Clark, 1996);

- INTENTIONAL INFLUENCES factors, inspired by the model of Jordan (2000b).

We develop features representing these factors, then use the features to represent examples of object descriptions and the context in which they occur for the purpose of learning a model of content selection for object descriptions.

Dale and Reiter's INCREMENTAL model focuses on the production of near-minimal subsequent references that allow the hearer to reliably distinguish the task object from similar task objects. Following Grosz and Sidner (1986), Dale and Reiter's algorithm utilizes discourse structure as an important factor in determining which objects the current object must be distinguished from. The model of Clark, Brennan and Wilkes-Gibbs is based on the notion of a CONCEPTUAL PACT, i.e. the conversants attempt to coordinate with one another by establishing a conceptual pact for describing an object. Jordan's INTENTIONAL INFLUENCES model is based on the assumption that the underlying communicative and task-related inferences are important factors in accounting for non-minimal descriptions. We describe these models in more detail in Section 3 and explain why we expect these models to work well in combination.

Many aspects of the underlying content selection models are not well-defined from an implementation point of view, so it may be necessary to experiment with different definitions and related parameter settings to determine which will produce the best performance for a model, as was done with the parameter setting experiments carried out by Jordan (2000b).[1] However, in the experiments we describe in this paper, we strive for feature representations that will allow the machine learner to take on more of the task of finding optimal settings and otherwise use the results reported by Jordan (2000b) for guidance. The only variation we test here is the representation of discourse structure for those models that require it. Otherwise, explicit tests of different interpretations of the models are left to future work.

We report on a set of experiments designed to establish the predictive power of the factors emphasized in the three models by using machine learning to train and test the content selection component of an object description generator on a set of 393 object descriptions from the corpus of COCONUT dialogues. The generator goes beyond each of the models' accounts for anaphoric expressions to address the more general problem of generating both initial and subsequent expressions. We provide the machine learner with distinct sets of features motivated by these models, in addition to discourse features motivated by assumed familiarity distinctions (Prince, 1981) (i.e. new vs. evoked vs. inferable discourse entities), and dialogue specific features such as the speaker of the object description, its absolute location in the discourse, and the problem that the conversants are currently trying to solve. We evaluate the object description generator by comparing its predictions against what humans said at the same point in the dialogue and only counting as correct those that exactly match the content of the human generated object descriptions (Oberlander, 1998).[2] This provides a rigorous test of the object description generator since in all likelihood there are other object descriptions that would have achieved the speaker's communicative goals.

---

1. Determining optimal parameter settings for a machine learning algorithm is a similar issue (Daelemans & Hoste, 2002) but at a different level. We use the same machine learner and parameter settings for all our experiments although searching for optimal machine learner parameter settings may be of value in further improving performance.

We also quantify the contribution of each feature set to the performance of the object description generator. The results indicate that the INTENTIONAL INFLUENCES features, the INCREMENTAL features and the CONCEPTUAL PACT features are all independently significantly better than the majority class baseline for this task, with the INTENTIONAL INFLUENCES model (42.4%) performing significantly better than either the INCREMENTAL model (30.4%) or the CONCEPTUAL PACT model (28.9%). However, the *best* performing models combine features from all the models, achieving accuracies at matching human performance near 60.0%, a large improvement over the majority class baseline of 17% in which the generator simply guesses the most frequent attribute combination. Surprisingly, our results in experimenting with different discourse structure parameter settings show that features derived from a simple recency-based model of discourse structure contribute as much to this particular task as one based on intentional structure.

The COCONUT dataset is small compared to those used in most machine learning experiments. Smaller datasets run a higher risk of overfitting and thus specific performance results should be interpreted with caution. In addition the COCONUT corpus represents only one type of dialogue; typed, collaborative, problem solving dialogues about constraint satisfaction problems. While the models and suggested features focus on general communicative issues, we expect variations in the task involved and in the communication setting to impact the predictive power of the feature sets. For example, the CONCEPTUAL PACT model was developed using dialogues that focus on identifying novel, abstract figures. Because the figures are abstract it is not clear at the start of a series of exercises what description will best help the dialogue partner identify the target figure. Thus the need to negotiate a description for the figures is more prominent than in other tasks. Likewise we expect constraint satisfaction problems and the need for joint agreement on a solution to cause the INTENTIONAL INFLUENCES model to be more prominent for the COCONUT dialogues. But the fact that the CONCEPTUAL PACT features show predictive power that is significantly better than the baseline suggests that while the prominence of each model inspired feature set may vary across tasks and communication settings, we expect each to have a significant contribution to make to a content selection model.

Clearly, for those of us whose ultimate goal is a general model of content selection for dialogue, we need to carry out experiments on a wide range of dialogue types. But for those of us whose ultimate goal is a dialogue application, one smaller corpus that is representative of the anticipated dialogues is probably preferable. Despite the two notes of caution we expect our feature representations to suggest a starting point for both larger endeavors.

---

2. Note that the more attributes a discourse entity has, the harder it is to achieve an exact match to a human description, i.e. for this problem the object description generator must correctly choose among 16 possibilities represented by the power set of the four attributes.

Previous research has applied machine learning to several problems in natural language generation, such as cue word selection (Di Eugenio, Moore, & Paolucci, 1997), accent placement (Hirschberg, 1993), determining the form of an object description (Poesio, 2000), content ordering (Malouf, 2000; Mellish, Knott, Oberlander, & O'Donnell, 1998; Duboue & McKeown, 2001; Ratnaparkhi, 2002), sentence planning (Walker, Rambow, & Rogati, 2002), re-use of textual descriptions in automatic summarization (Radev, 1998), and surface realization (Langkilde & Knight, 1998; Bangalore & Rambow, 2000; Varges & Mellish, 2001).

The only other machine learning approaches for content selection are those of Oh and Rudnicky (2002) and of Roy (2002). Oh and Rudnicky report results for automatically training a module for the CMU Communicator system that selects the attributes that the system should express when *implicitly confirming* flight information in an ongoing dialogue. For example, if the caller said *I want to go to Denver on Sunday*, the implicit confirmation by the system might be *Flying to Denver on Sunday*. They experimentally compared a statistical approach based on bigram models with a strategy that only confirms information that the system has just heard for the first time, and found that the two systems performed equally well. Roy reports results for a spoken language generator that is trained to generate visual descriptions of geometric objects when provided with features of visual scenes. Roy's results show that the understandability of the automatically generated descriptions is only 8.5% lower than human-generated descriptions. Unlike our approach, neither of these consider the effects of ongoing dialogue with a dialogue partner, or the effect of the dialogue context on the generated descriptions. Our work, and the theoretical models it is based on, explicitly focus on the processes involved in generating *descriptions* and *redescriptions* of objects in interactive dialogue that allow the dialogue partners to remain aligned as the dialogue progresses (Pickering & Garrod, 2004).

The most relevant prior work is that of Jordan (2000b). Jordan implemented Dale and Reiter's INCREMENTAL model and developed and implemented the INTENTIONAL INFLU-ENCES model, which incorporates the INCREMENTAL model, and tested them both against the COCONUT corpus. Jordan also experimented with different parameter settings for vague parts of the models. The results of this work are not directly comparable because Jordan only tested rules for subsequent reference, while here we attempt to learn rules for generating both initial and subsequent references. However, using a purely rule-based approach, the best accuracy that Jordan reported was 69.6% using a non-stringent scoring criterion (not an exact match) and 24.7% using the same stringent exact match scoring used here. In this paper, using features derived from Jordan's corpus annotations, and applying rule induction to induce rules from training data, we achieve an exact match accuracy of nearly 47% when comparing to the most similar model and an accuracy of nearly 60% when comparing to the best overall model. These results appear to be an improvement over those reported by Jordan (2000b), given both the increased accuracy and the ability to generate initial as well as subsequent references.

Section 2 describes the COCONUT corpus, definitions of discourse entities and object descriptions for the COCONUT domain, and the annotations on the corpus that we use to derive the feature sets. Section 3 presents the theoretical models of content selection for object descriptions in more detail and describes the features inspired by the models. Section 4 describes the experimental design and Section 5 presents the quantitative results of testing the learned rules against the corpus, discusses the features that the machine learner identifies as important, and provides examples of the rules that are learned. Section 6 summarizes the results and discusses future work.

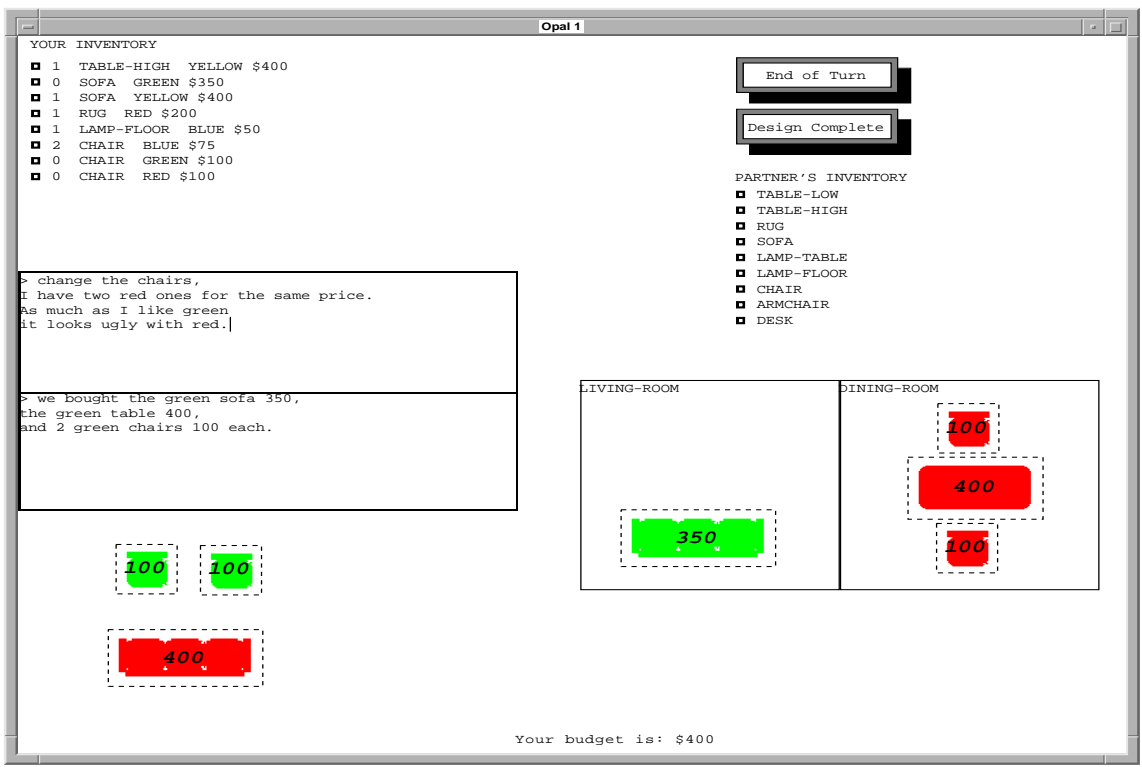

Figure 2: A snapshot of the interface for the COCONUT task

## 2. The COCONUT Corpus

The COCONUT corpus is a set of 24 computer-mediated dialogues consisting of a total of 1102 utterances. The dialogues were collected in an experiment where two human subjects collaborated on a simple design task, that of buying furniture for two rooms of a house (Di Eugenio et al., 2000). Their collaboration was carried out through a typed dialogue in a workspace where each action and utterance was automatically logged. An excerpt of a COCONUT dialogue is in Figure 1. A snapshot of the workspace for the COCONUT experiments is in Figure 2.

In the experimental dialogues, the participants' main goal is to negotiate the purchases; the items of highest priority are a sofa for the living room and a table and four chairs for the dining room. The participants also have specific secondary goals which further constrain the problem solving task. Participants are instructed to try to meet as many of these goals as possible, and are motivated to do so by rewards associated with satisfied goals.

The secondary goals are: 1) match colors within a room, 2) buy as much furniture as you can, 3) spend all your money. The participants are told which rewards are associated with achieving each goal.

Each participant is given a separate budget (as shown in the mid-bottom section of Figure 2) and an inventory of furniture (as shown in the upper-left section of Figure 2). Furniture types include sofas, chairs, rugs and lamps, and the possible colors are red, green, yellow or blue. Neither participant knows what is in the other's inventory or how much money the other has. By sharing information during the conversation, they can combine their budgets and select furniture from each other's inventories. Note that since a participant does not know what furniture his partner has available until told, there is a menu (see the mid-right section of Figure 2) that allows the participant to create furniture items based on his partner's description of the items available. The participants are equals and purchasing decisions are joint. In the experiment, each set of participants solved one to three scenarios with varying inventories and budgets. The problem scenarios varied task complexity by ranging from tasks where items are inexpensive and the budget is relatively large, to tasks where the items are expensive and the budget relatively small.

## 2.1 Discourse Entities and Object Descriptions in the Corpus

A *discourse model* is used to keep track of the objects discussed in a discourse. As an object is described, the conversants relate the information about the object in the utterance to the appropriate mental representation of the object in the discourse model (Karttunen, 1976; Webber, 1978; Heim, 1983; Kamp & Reyle, 1993; Passonneau, 1996). The model contains discourse entities, attributes and links between entities (Prince, 1981). A *discourse entity* is a variable or placeholder that indexes the information about an object described in a particular linguistic description to an appropriate mental representation of the object. The discourse model changes as the discourse progresses. When an object is first described, a discourse entity such as $e_i$ is added to the discourse model. As new utterances are produced, additional discourse entities may be added to the model when new objects are described, and new attributes may get associated with $e_i$ whenever it is redescribed. Attributes are not always supplied by a noun phrase (NP). They may arise from other parts of the utterance or from discourse inference relations that link to other discourse entities.

To illustrate the discourse inference relations relevant to COCONUT, in (1b) *the green set* is an example of a new discourse entity which has a set/subset discourse inference relation to the three distinct discourse entities for *2 \$25 green chairs*, *2 \$100 green chairs* and *1 \$200 green table*.

(1) a. : I have [2 \$25 green chairs] and [a \$200 green table].

b. : I have [2 \$100 green chairs]. Let's get [the green set].

A class inference relation exists when the referent of a discourse entity has a subsumption relationship with a previous discourse entity. For example, in (2) *the table* and *your green one* have a subsumption relationship.

(2) Let's decide on [the table] for the dining room. How about [your green one]?

A common noun anaphora inference relation occurs in the cases of one anaphora and null anaphora. For example, in (3) each of the marked NPs in the last part of the utterance has a null anaphora relation to the marked NP in the first part. Note that this example also has a class inference relation as well.

(3)     I have [a variety of high tables] ,[green], [red] and [yellow] for 400, 300, and 200.

Discourse entities can also be related by predicative relationships such as *is*. For example, in (4) the entities defined by *my cheapest table* and *a blue one for $200* are not the same discourse entities but the information about one provides more information about the other. Note that this example also includes common noun anaphora and class inference relations.

(4)     [My cheapest table] is [a blue one for $200].

An object description is any linguistic expression (usually an NP) that initiates the creation or update of a discourse entity for a furniture item, along with any explicit attributes expressed within the utterance. We consider the attributes that are explicitly expressed outside of an NP to be part of the object description since they can be realized either as part of the noun phrase that triggers the discourse entity or elsewhere in the utterance. Attributes that are inferred (e.g. quantity from "a" or "the") help populate the discourse entity but are not considered part of an object description since inferred attributes may or may not reflect an explicit choice. The inferred attribute could be a side-effect of the surface structure selected for realizing the object description.[3]

## 2.2 Corpus Annotations

After the corpus was collected, it was annotated by human coders for three types of features: PROBLEM-SOLVING UTTERANCE LEVEL features as shown in Figure 3, DISCOURSE UTTERANCE LEVEL features as illustrated in Figure 4 and DISCOURSE ENTITY LEVEL features as illustrated in Figure 5. Some additional features are shown in Figure 6. Each of the feature encodings shown are for the dialogue excerpt in Figure 1.

All of the features were hand-labelled on the corpus because it is a human-human corpus but, as we will discuss further at the end of this section, many of these features would need to be established by a system for its collaborative problem solving component to function properly.

Looking first at Figure 6, it is the explicit attributes (as described in the previous section) that are to be predicted by the models we are building and testing. The remaining features are available as context for making the predictions.

The PROBLEM-SOLVING UTTERANCE LEVEL features in Figure 3 capture the problem solving state in terms of the goals and actions that are being discussed by the conversants, constraint changes that are implicitly assumed, or explicitly stated by the conversants, and the size of the solution set for the current constraint equations. The solution set size for a constraint equation is characterized as being *determinate* if the set of values is closed and represents that the conversants have shared relevant values with one another. An *indeterminate* size means that the set of values in still open and so a solution cannot yet be determined. The problem-solving features capture some of the situational or problem-solving influences that may effect descriptions and indicate the task structure from which the discourse structure can be derived (Terken, 1985; Grosz & Sidner, 1986). Each domain goal provides a discourse segment purpose so that each utterance that relates to a different domain goal or set of domain goals defines a new segment.

---

3. While the same is true of some of the attributes that are explicitly expressed (e.g. "I" in subject position expresses the ownership attribute), most of the attribute types of interest in the corpus are adjuncts (e.g. "Let's buy the chair [for $100].").

| Utterance | Goal/ Action Label | Introduce or continue | Goal/ Action Identifier | Change in Constraints | Solution Size |
|---|---|---|---|---|---|
| 37 | SelectOptionalItemLR | introduce | act4 | drop color match | indeterminate |
| 38 | SelectOptionalItem | introduce | act5 | color,price limit | indeterminate |
| 39 | SelectOptionalItem | continue | act5 | none | indeterminate |
| 40 | SelectOptionalItemLR | continue | act4 | none | determinate |
| 42 | SelectOptionalItemLR | continue | act4 | none | determinate |
| 43 | SelectOptionalItemDR | continue | act5 | none | indeterminate |
| 44 | SelectOptionalItemDR | continue | act5 | none | determinate |
| 46 | SelectOptionalItemDR | continue | act5 | none | determinate |
| 47 | SelectOptionalItemDR | continue | act4 | none | determinate |
|  | SelectOptionalItemLR | continue | act5 |  |  |
| 48 | SelectOptionalItemDR | continue | act5 | none | determinate |
| 51 | SelectOptionalItemDR, | continue | act5, | none | determinate |
|  | SelectChairs | introduce | act3 |  |  |
| 52 | SelectOptionalItemLR | continue | act4 | none | determinate |

Figure 3: Problem solving utterance level annotations for utterances relevant to problem solving goals and actions for the dialogue excerpt in Figure 1

| Utterance | Influence on Listener | Influence on Speaker |
|---|---|---|
| 37 | ActionDirective | Offer |
| 40 | ActionDirective | Commit |
| 42 | ActionDirective | Commit |
| 43 | OpenOption | nil |
| 44 | ActionDirective | Offer |
| 46 | ActionDirective | Commit |
| 47 | ActionDirective | Commit |
| 48 | ActionDirective | Commit |
| 49 | ActionDirective | Offer |
| 51 | ActionDirective | Offer |
| 52 | ActionDirective | Commit |

Figure 4: Discourse utterance level annotations for utterances relevant to establishing joint agreements for the dialogue excerpt in Figure 1

| Utterance | Reference and Coreference | Discourse Inference Relations | Attribute Values | Argument for Goal/Action Identifier |
|---|---|---|---|---|
| 37 | initial ref-1 | nil | my,1,yellow,rug,150 | act4 |
| 38 | initial ref-2 | nil | your,furniture,100 | act5 |
| 39 | initial ref-3 | class to ref-20 | my,furniture,100 | act5 |
| 40 | corefers ref-1 | nil | your,1,yellow,rug,150 | act4 |
| 42 | corefers ref-1 | nil | my,1,rug,150 | act4 |
| 43 | initial ref-4 | nil | my,1,green,chair | act5 |
| 44 | corefers ref-4 | CNAnaphora ref-4 | my,100 | act5 |
| 47 | corefers ref-1 | nil | your,1,yellow,rug | act4 |
| 47 | corefers ref-4 | nil | your,1,green,chair | act5 |
| 48 | corefers ref-4 | nil | my,1,green,chair,100 | act5 |
| 51 | corefers ref-4 | nil | 1,green,chair | act5 |
| 51 | initial ref-5 | set of ref-12,ref-16 | chair | act3 |
| 52 | corefers ref-1 | | 1,yellow rug | act4 |

Figure 5: Discourse entity level annotations for utterances referring to furniture items in Figure 1

| Utterance | Speaker | Explicit Attributes | Inferred Attributes | Description |
|---|---|---|---|---|
| 37 | G | type,color,price,owner | quantity | a yellow rug for 150 dollars |
| 38 | G | type,color,price,owner | | furniture ... for 100 dollars |
| 39 | S | type,price,owner | | furniture ... 100 dollars |
| 40 | S | type,color,price,owner | quantity | the yellow rug for $150 |
| 42 | G | type,price,owner | quantity | the rug for 150 dollars |
| 43 | G | type,color,owner | quantity | a green chair |
| 44 | G | price,owner | | [0] for 100 dollars |
| 47 | S | type,color | owner,quantity | the yellow rug |
| 47 | S | type,color | owner,quantity | the green chair |
| 48 | G | type,color,price,owner | quantity | the green 100 dollar chair |
| 51 | S | type,color | quantity | the green chair |
| 51 | S | type | quantity | the other chairs |
| 52 | S | type,color | quantity | the yellow rug |

Figure 6: Additional features for the dialogue excerpt in Figure 1

The DISCOURSE UTTERANCE LEVEL features in Figure 4 encode the influence the utterance is expected to have on the speaker and the listener as defined by the DAMSL scheme (Allen & Core, 1997). These annotations also help capture some of the situational influences that may effect descriptions. The possible influences on listeners include *open options*, *action directives* and *information requests*. The possible influences on speakers are *offers* and *commitments*. Open options are options that a speaker presents for the hearer's future actions, whereas with an action directive a speaker is trying to put a hearer under an obligation to act. There is no intent to put the hearer under obligation to act with an open option because the speaker may not have given the hearer enough information to act or the speaker may have clearly indicated that he does not endorse the action. Offers and commitments are both needed to arrive at a joint commitment to a proposed action. With an offer the speaker is conditionally committing to the action whereas with a commit the speaker is unconditionally committing. With a commit, the hearer may have already conditionally committed to the action under discussion, or the speaker may not care if the hearer is also committed to the action he intends to do.

The DISCOURSE ENTITY LEVEL features in Figure 5 define the discourse entities that are in the discourse model. Discourse entities, links to earlier discourse entities and the attributes expressed previously for a discourse entity at the NP-level and utterance level are inputs for an object description generator. Part of what is used to define the discourse entities is discourse reference relations which include initial, coreference and discourse inference relations between different entities such as the links we described earlier; set/subset, class, common noun anaphora and predicative. In addition, in order to link the expression to appropriate problem solving actions, the action for which the entity is an argument is also annotated. In order to test whether an acceptable object description is generated by a model for a discourse entity in context, the explicit attributes used to describe the entity are also annotated (recall Figure 6).

Which action an entity is related to helps associate entities with the correct parts of the discourse structure and helps determine which problem-solving situations are relevant to a particular entity. From the other discourse entity level annotations, initial representations of discourse entities and updates to them can be derived. For example, the initial representation for "I have a yellow rug. It costs $150." would include type, quantity, color and owner following the first utterance. Only the quantity attribute is inferred. After the second utterance the entity would be updated to include price.

The encoded features all have good inter-coder reliability as shown by the KAPPA values given in Table 1 (Di Eugenio et al., 2000; Jordan, 2000b; Krippendorf, 1980). These values are all statistically significant for the size of the labelled data set, as shown by the p-values in the table.

| Discourse Entity Level | Reference and Coreference | Discourse Inference Relations | Argument for Goal/ Action | Attributes | |
|---|---|---|---|---|---|
| | .863 (z=19, p<.01) | .819 (z=14, p<.01) | .857 (z=16, p<.01) | .861 (z=53, p<.01) | |
| Problem Solving | Introduce Goal/Action | Continue Goal/Action | Change in Constraints | Solution Size | Goal/Action |
| Utterance Level | .897 (z=8, p<.01) | .857 (z=27, p<.01) | .881 (z=11, p<.01) | .8 (z=6, p<.01) | .74 (z=12, p<.01) |
| Discourse Utterance Level | Influence on Listener | Influence on Speaker | | | |
| | .72 (z=19, p<.01) | .72 (z=13, p<.01) | | | |

Table 1: Kappa values for the annotation scheme

While the availability of some of this annotated information in a dialogue system is currently an ongoing challenge for today's systems, a system that is to be a successful dialogue partner in a collaborative problem solving dialogue, where all the options are not known a priori, will have to model and update discourse entities, understand the current problem solving state and what has been agreed upon, and be able to make, accept or reject proposed solutions. Certainly, not all dialogue system domains and communicative settings will need all of this information and likewise some of the information that is essential for other domains and settings will not be necessary to engage in a COCONUT dialogue.

The experimental data consists of 393 non-pronominal object descriptions from 13 dialogues of the COCONUT corpus as well as features constructed from the annotations described above. The next section explains in more detail how the annotations are used to construct the features used in training the models.

## 3. Representing Models of Content Selection for Object Descriptions as Features

In Section 1, we described how we would use the annotations on the COCONUT corpus to construct feature sets motivated by theories of content selection for object descriptions. Here we describe these theories in more detail, and present, with each theory, the feature sets that are inspired by the theory. In Section 4 we explain how these features are used to automatically learn a model of content selection for object descriptions. In order to be used in this way, all of the features must be represented by continuous (numeric), set-valued, or symbolic (categorial) values.

Models of content selection for object descriptions attempt to explain what motivates a speaker to use a particular set of attributes to describe an object, both on the first mention of an object as well as in subsequent mentions. In an extended discourse, speakers often redescribe objects that were introduced earlier in order to say something more about the object or the event in which it participates. We will test in part an assumption that many of the factors relevant for redescriptions will also be relevant for initial descriptions.

All of the models described below have previously had rule-based implementations of them tested on the COCONUT corpus and were all found to be nearly equally good at explaining the redescriptions in the corpus (Jordan, 2000b). All of them share a basic assumption about the speaker's goal when redescribing a discourse entity already introduced into the discourse model in prior conversation. The speaker's primary goal is *identification*, i.e. to generate a linguistic expression that will efficiently and effectively re-evoke the appropriate discourse entity in the hearer's mind. A redescription must be *adequate* for re-evoking the entity unambiguously, and it must do so in an *efficient* way (Dale & Reiter, 1995). One factor that has a major effect on the adequacy of a redescription is the fact that a discourse entity to be described must be distinguished from other discourse entities in the discourse model that are currently salient. These other discourse entities are called *distractors*. Characteristics of the discourse entities evoked by the dialogue such as recency and frequency of mention, relationship to the task goals, and position relative to the structure of the discourse are hypothesized as means of determining which entities are mutually salient for both conversants.

---

- what is mutually known: type-mk, color-mk, owner-mk, price-mk, quantity-mk
- reference-relation: one of `initial`, `coref`, `set`, `class`, `cnanaphora`, `predicative`

Figure 7: Assumed Familiarity Feature Set.

---

We begin the encoding of features for the object description generator with features representing the fundamental aspects of a discourse entity in a discourse model. We divide these features into two sets: the ASSUMED FAMILIARITY feature set and the INHERENT feature set. The ASSUMED FAMILIARITY features in Figure 7 encode all the information about a discourse entity that is already represented in the discourse model at the point in the discourse at which the entity is to be described. These attributes are assumed to be mutually known by the conversational participants and are represented by five boolean features: *type-mk, color-mk, owner-mk, price-mk, quantity-mk*. For example, if *type-mk* has the value of *yes*, this represents that the type attribute of the entity to be described is mutually known.

Figure 7 also enumerates a *reference-relation* feature as described in Section 2 to encode whether the entity is new (`initial`), evoked (`coref`) or inferred relative to the discourse context. The types of inferences supported by the annotation are set/subset, class, common noun anaphora (e.g. one and null anaphora), and predicative (Jordan, 2000b), which are represented by the values (`set,class,cnanaphora,predicative`). These reference relations are relevant to both initial and subsequent descriptions.

---

- utterance-number, speaker-pair, speaker, problem-number
- attribute values:

  - type: one of `sofa, chair, table, rug, lamp, superordinate`
  - color: one of `red, blue, green, yellow`
  - owner: one of `self, other, ours`
  - price: range from $50 to $600
  - quantity: range from 0 to 4.

Figure 8: INHERENT Feature Set: Task, Speaker and Discourse Entity Specific features.

---

The INHERENT FEATURES in Figure 8 are a specific encoding of particulars about the discourse situation, such as the speaker, the task, and the actual values of the entity's known attributes (*type, color, owner, price, quantity*). We supply the values for the attributes in case there are preferences associated with particular values. For example, there may be a preference to include quantity, when describing a set of chairs, or price, when it is high.

The inherent features allow us to examine whether there are individual differences in selection models (*speaker, speaker-pair*), or whether specifics about the attributes of the object, the location within the dialogue (*utterance-number*), and the problem difficulty (*problem-number*) play significant roles in selecting attributes. The attribute values for an entity are derived from annotated attribute features and the reference relations.

We don't expect rules involving this feature set to generalize well to other dialogue situations. Instead we expect them to lead to a SITUATION SPECIFIC model. Whenever these features are used there is overfitting regardless of the training set size. Consider that a particular speaker, speaker-pair or utterance number are specific to particular dialogues and are unlikely to occur in another dialogue, even a new COCONUT dialogue. These feature representations would have to be abstracted to be of value in a generator.

## 3.1 Dale and Reiter's INCREMENTAL Model

Most computational work on generating object descriptions for subsequent reference (Appelt, 1985a; Kronfeld, 1986; Reiter, 1990; Dale, 1992; Heeman & Hirst, 1995; Lochbaum, 1995; Passonneau, 1996; van Deemter, 2002; Gardent, 2002; Krahmer, van Erk, & Verleg, 2003) concentrates on how to produce a minimally complex expression that singles out the discourse entity from a set of distractors. The set of contextually salient distractors is identified via a model of discourse structure as mentioned above. Dale and Reiter's INCREMENTAL model is the basis of much of the current work that relies on discourse structure to determine the content of object descriptions for subsequent reference.

The most commonly used account of discourse structure for task-oriented dialogues is Grosz and Sidner's (1986) theory of the attentional and intentional structure of discourse. In this theory, a data structure called a focus space keeps track of the discourse entities that are salient in a particular context, and a stack of focus spaces is used to store the focus spaces for the discourse as a whole. The content of a focus space and operations on the stack of focus spaces is determined by the structure of the task. A change in task or topic indicates the start of a new discourse segment and a corresponding focus space. All of the discourse entities described in a discourse segment are classified as salient for the dialogue participants while the corresponding focus space is on the focus stack. Approaches that use a notion of discourse structure take advantage of this representation to produce descriptors that are minimally complex given the current focus space, i.e. the description does not have to be unambiguous with respect to the global discourse.

According to Dale and Reiter's model, a descriptor containing information that is not needed to identify the referent given the current focus space would not be minimally complex but a small number of overspecifications that appear relative to the identification goal are expected and can be explained as artifacts of cognitive processing limits. Trying to produce a minimally complex description can be seen as an implementation of the two parts of Grice's *Maxim of Quantity*, according to which an utterance should both say as much as is required, and no more than is required (Grice, 1975). Given an entity to describe and a distractor set defined by the entities in the current focus space, the INCREMENTAL model incrementally builds a description by checking a static ordering of attribute types and selecting an attribute to include in the description if and only if it eliminates some of the remaining distractors. As distractors are ruled out, they no longer influence the selection process.

- Distractor Frequencies: type-distractors, color-distractors, owner-distractors, price-distractors, quantity-distractors

- Attribute Saliency: majority-type, majority-type-freq, majority-color, majority-color-freq, majority-price, majority-price-freq, majority-owner, majority-owner-freq, majority-quantity, majority-quantity-freq

Figure 9: CONTRAST SET Feature Sets

A set of features called CONTRAST SET features are used to represent aspects of Dale and Reiter's model. See Figure 9. The goal of the encoding is to represent whether there are distractors present in the focus space which might motivate the inclusion of a particular attribute. First, the *distractor frequencies* encode how many distractors have an attribute value that is different from that of the entity to be described.

The INCREMENTAL model also utilizes a preferred salience ordering for attributes and eliminates distractors as attributes are added to a description. For example, adding the attribute *type* when the object is a chair, eliminates any distractors that aren't chairs. A feature based encoding cannot easily represent a distractor set that changes as attribute choices are made. To compensate, our encoding treats attributes instead of objects as distractors so that the *attribute saliency* features encode which attribute values are most salient for each attribute type, and a count of the number of distractors with this attribute value. For example, if 5 of 8 distractors are **red** then *majority-color* is **red** and the *majority-color-freq* is 5. Taking the view of attributes as distractors has the advantage that the preferred ordering of attributes can adjust according to the focus space. This interpretation of Dale and Reiter's model was shown to be statistically similar to the strict model but with a higher mean match to the corpus (Jordan, 2000b). Thus our goal in adding these additional features is to try to obtain the best possible performance for the INCREMENTAL model.

Finally, an open issue with deriving the distractors is how to define a focus space (Walker, 1996a). As described above, Grosz and Sidner's theory of discourse creates a data structure called a focus space for each discourse segment, where discourse segments are based on the intentions underlying the dialogue. However Grosz and Sidner provide no clear criterion for assigning the segmentation structure. In order to explore what definition variations will work best, we experiment with three focus space definitions, two very simple focus space definitions based on recency, and the other based on intentional structure as described below. To train and test for the three focus space definitions, we create separate datasets for each of the three. To our knowledge, this is the first empirical comparison of Grosz and Sidner's model with a simpler model for any discourse-related task.

For intentional structure, we utilize the problem solving utterance features hand-labelled on the COCONUT corpus with high reliability as discussed above in Section 2. The annotated task goals are used to derive an intentional structure for the discourse, which provides a segmentation of the discourse, as described by Grosz and Sidner (1986). The current focus space as defined by the annotated task goals is used to define segment distractors. This dataset we label as SEGMENT. For recency, one extremely simple focus space definition uses only the discourse entities from the most recent utterance as possible distractors. This dataset we label as ONE UTTERANCE. The second extremely simple focus space definition only considers the discourse entities from the last five utterances as possible distractors. This dataset we label as FIVE UTTERANCE. For each dataset, the features in Figure 9 are computed relative to the distractors determined by its focus space definition.

## 3.2 Jordan's INTENTIONAL INFLUENCES Model

Jordan (2000b) proposed a model to select attributes for object descriptions for subsequent reference called the INTENTIONAL INFLUENCES model. This model posits that along with the identification goal, task-related inferences and the agreement process for task negotiation are important factors in selecting attributes. Attributes that are not necessary for identification purposes may be intentional redundancies with a communicative purpose (Walker, 1996b) and not always just due to cognitive processing limits on finding minimally complex descriptions (Jordan, 2000b).

A goal-directed view of sentence generation suggests that speakers can attempt to satisfy multiple goals with each utterance (Appelt, 1985b). It suggests that this strategy also applies to lower-level forms within the utterance (Stone & Webber, 1998). That is, the same form can opportunistically contribute to the satisfaction of multiple goals. This many-one mapping of goals to linguistic forms is more generally referred to as *overloading intentions* (Pollack, 1991). Subsequent work has shown that this overloading can involve trade-offs across linguistic levels. That is, an intention which is achieved by complicating a form at one level may allow the speaker to simplify another level by omitting important information. For example, a choice of clausal connectives at the pragmatic level can simplify the syntactic level (Di Eugenio & Webber, 1996), and there are trade-offs in word choice at the syntax and semantics levels (Stone & Webber, 1998).

The INTENTIONAL INFLUENCES model incorporates multiple communicative and problem solving goals in addition to the main identification goal in which the speaker intends the hearer to re-evoke a particular discourse entity. The contribution of this model is that it overloads multiple, general communicative and problem solving goals when generating a description. When the model was tested on the COCONUT corpus, inferences about changes in the problem solving constraints, about conditional and unconditional commitments to proposals, and about the closing of goals were all shown to be relevant influences on attribute selection (Jordan, 2000a, 2002) while goals to verify understanding and infer informational relations were not (Jordan, 2000b).[4]

The features used to approximate Jordan's model are in Figure 10. These features cover all of the general communicative and problem solving goals hypothesized by the model except for the identification goal and the information relation goal. Because of the difficulty of modelling an information relation goal with features, its representation is left to future work.[5]

---

4. A different subset of the general goals covered by the model are expected to be influential for other domains and communication settings, therefore a general object description generator would need to be trained on a wide range of corpora.

5. Information relation goals may relate two arbitrarily distant utterances and additional details beyond distance are expected to be important. Because this goal previously did not appear relevant for the COCONUT corpus (Jordan, 2000b), we gave it a low priority for implementation.

- task situation: goal, colormatch, colormatch-constraintpresence, pricelimit, pricelimit-constraintpresence, priceevaluator, priceevaluator-constraintpresence, colorlimit, colorlimit-constraintpresence, priceupperlimit, priceupperlimit-constraintpresence

- agreement state: influence-on-listener, commit-speaker, solution-size, prev-influence-on-listener, prev-commit-speaker, prev-solution-size, distance-of-last-state-in-utterances, distance-of-last-state-in-turns, ref-made-in-prev-action-state, speaker-of-last-state, prev-ref-state

- previous agreement state description: prev-state-type-expressed, prev-state-color-expressed, prev-state-owner-expressed, prev-state-price-expressed, prev-state-quantity-expressed

- solution interactions: color-contrast, price-contrast

Figure 10: Intentional Influences Feature Set.

The task situation features encode inferable changes in the task situation that are related to item attributes, where *colormatch* is a boolean feature that indicates whether there has been a change in the color match constraint. The *pricelimit*, *colorlimit* and *priceupperlimit* features are also boolean features representing that there has been a constraint change related to setting limits on values for the price and color attributes. The features with *constraintpresence* appended to a constraint feature name are symbolic features that indicate whether the constraint change was implicit or explicit. For example, if there is an agreed upon constraint to try to select items with the same color value for a room, and a speaker wants to relax that constraint then the feature *colormatch* would have the value `yes`. If the speaker communicated this explicitly by saying "Let's forget trying to match colors." then the *constraintpresence* feature would have the value `explicit` and otherwise it would have the value `implicit`. If the constraint change is not explicitly communicated and the speaker decides to include a color attribute when it is not necessary for identification purposes, it may be to help the hearer infer that he means to drop the constraint

The agreement state features in Figure 10 encode critical points of agreement during problem solving. Critical agreement states are (Di Eugenio et al., 2000):

- propose: the speaker *offers* the entity and this conditional commitment results in a *determinate solution size*.

- partner decidable option: the speaker *offers* the entity and this conditional commitment results in an *indeterminate solution size*.

- unconditional commit: the speaker *commits* to an entity.

- unendorsed option: the speaker *offers* the entity but does not show any *commitment* to using it when the *solution size is already determinate*.

For example, if a dialogue participant is *unconditionally committing* in response to a *proposal*, she may want to verify that she has the same item and the same entity description as her partner by repeating back the previous description. The features that encode these critical agreement states include some DAMSL features (*influence-on-listener,* *commit-speaker, prev-influence-on-listener, prev-commit-speaker*), progress toward a solution (*solution-size, prev-solution-size, ref-made-in-prev-action-state*), and features inherent to an agreement state (*speaker-of-last-state, distance-of-last-state-in-utterances, distance-of-last-state-in-turns*). The features that make reference to a state are derived from the agreement state features and a more extensive discourse history than can be encoded within the feature representation. In addition, since the current agreement state depends in part on the previous agreement state, we added the derived agreement state. The previous agreement state description features in Figure 10 are booleans that capture dependencies of the model on the content of the description from a previous state. For example, if the previous agreement state for an entity expressed only type and color attributes then this would be encoded *yes* for prev-state-type-expressed and prev-state-color-expressed and *no* for the rest.

The solution interactions features in Figure 10 represent situations where multiple proposals are under consideration which may contrast with one another in terms of solving color-matching goals (*color-contrast*) or price related goals (*price-contrast*). When the boolean feature *color-contrast* is true, it means that the entity's color matches with the partial solution that has already been agreed upon and contrasts with the alternatives that have been proposed. In this situation, there may be grounds for endorsing this entity relative to the alternatives. For example, in response to S's utterance [37] in Figure 1, in a context where G earlier introduced one blue rug for $175, G could have said "Let's use my blue rug." in response. In this case the blue rug would have a true value for *color-contrast* because it has a different color than the alternative, and it matches the blue sofa that had already been selected.

The boolean feature *price-contrast* describes two different situations. When the feature *price-contrast* is true, it either means that the entity has the best price relative to the alternatives, or when the problem is nearly complete, that the entity is more expensive than the alternatives. In the first case, the grounds for endorsement are that the item is cheaper. In the second case, it may be that the item will spend out the remaining budget which will result in a higher score for the problem solution.

Note that although the solution interaction features depend upon the agreement states, in that it is necessary to recognize proposals and commitments in order to identify alternatives and track agreed upon solutions, it is difficult to encode such extensive historical information directly in a feature representation. Therefore the solution interaction features are derived, and the derivation includes heuristics that use agreement state features for estimating partial solutions. A sample encoding for the dialogue excerpt in Figure 1 for its problem solving utterance level annotations and agreement states were given in Figures 3 and 4.

## 3.3 Brennan and Clark's CONCEPTUAL PACT Model

Brennan and Clark's CONCEPTUAL PACT model focuses on the bidirectional adaptation of each conversational partner to the linguistic choices of the other conversational participant. The CONCEPTUAL PACT model suggests that dialogue participants negotiate a description that both find adequate for describing an object (Clark & Wilkes-Gibbs, 1986; Brennan & Clark, 1996). The speaker generates trial descriptions that the hearer modifies based on which object he thinks he is suppose to identify. The negotiation continues until the participants are confident that the hearer has correctly identified the intended object.

Brennan and Clark (1996) further point out that lexical availability, perceptual salience and a tendency for people to reuse the same terms when describing the same object in a conversation, all significantly shape the descriptions that people generate. These factors may then override the informativeness constraints imposed by Grice's Quantity Maxim. Lexical availability depends on how an object is best conceptualized and the label associated with that conceptualization (e.g. is the referent "an item of furniture" or "a sofa"). With perceptual salience, speakers may include a highly salient attribute rather than just the attributes that distinguish it from its distractors, e.g. "the $50 red sofa" when "the $50 sofa" may be informative enough. Adaptation to one's conversational partner should lead to a tendency to reuse a previous description.

The tendency to reuse a description derives from a combination of the most recent, successfully understood description of the object, and how often the description has been used in a particular conversation. However, this tendency is moderated by the need to adapt a description to changing problem-solving circumstances and to make those repeated descriptions even more efficient as their precedents become more established for a particular pairing of conversational partners. Recency and frequency effects on reuse are reflections of a coordination process between conversational partners in which they are negotiating a shared way of labelling or *conceptualizing* the referent. Different descriptions may be tried until the participants agree on a conceptualization. A change in the problem situation may cause the conceptualization to be embellished with additional attributes or may instigate the negotiation of a new conceptualization for the same referent.

The additional features suggested by this model include the previous description since that is a candidate conceptual pact, how long ago the description was made, and how frequently it was referenced. If the description was used further back in the dialogue or was referenced frequently, that could indicate that the negotiation process had been completed. Furthermore, the model suggests that, once a pact has been reached, that the dialogue participants will continue to use the description that they previously negotiated unless the problem situation changes. The continued usage aspect of the model is also similar to Passonneau's LEXICAL FOCUS model (Passonneau, 1995).

---

- interactions with other discourse entities: distance-last-ref, distance-last-ref-in-turns, number-prev-mentions, speaker-of-last-ref, distance-last-related

- previous description: color-in-last-exp, type-in-last-exp, owner-in-last-exp, price-in-last-exp, quantity-in-last-exp, type-in-last-turn, color-in-last-turn, owner-in-last-turn, price-in-last-turn, quantity-in-last-turn, initial-in-last-turn

- frequency of attributes: freq-type-expressed, freq-color-expressed, freq-price-expressed, freq-owner-expressed, freq-quantity-expressed

- stability history: cp-given-last-2, cp-given-last-3

Figure 11: CONCEPTUAL PACT Feature Set.

---

The CONCEPTUAL PACT features in Figure 11 encode how the current description relates to previous descriptions of the same entity. We encode recency information: when the entity was last described in terms of number of utterances and turns (*distance-last-ref, distance-last-in-turns*), when the last related description (e.g. set, class) was (*distance-last-related*), how frequently it was described (*number-prev-mentions*), who last described it (*speaker-of-last-ref*), and how it was last described in terms of turn and expression since the description may have been broken into several utterances (*color-in-last-exp, type-in-last-exp, owner-in-last-exp, price-in-last-exp, quantity-in-last-exp, type-in-last-turn, color-in-last-turn, owner-in-last-turn, price-in-last-turn, quantity-in-last-turn, initial-in-last-turn*). We also encode frequency information: the frequency with which attributes were expressed in previous descriptions of it (*freq-type-expressed, freq-color-expressed, freq-price-expressed, freq-owner-expressed, freq-quantity-expressed*), and a history of possible conceptual pacts that may have been formed; the attribute types used to describe it in the last two and last three descriptions of it if they were consistent across usages (*cp-given-last-2, cp-given-last-3*).

## 4. Experimental Method

The experiments utilize the rule learning program RIPPER (Cohen, 1996) to learn the content selection component of an object description generator from the object descriptions in the COCONUT corpus. Although any categorization algorithm could be applied to this problem given the current formulation, RIPPER is a good match for this particular setup because the if-then rules that are used to express the learned model can be used to easily compared with the theoretical models of content selection described above. One drawback is that RIPPER does not automatically take context into account during training so the discourse context must be represented via features as well. Although it might seem desirable to use RIPPER's own previous predictions as additional context during training, since it will consider them in practice, it is unnecessary and irrelevant to do so. The learned model will consist of generation rules that are relative to what is in the discourse as encoded features (i.e. what was actually said in the corpus) and any corrections it learns are only good for improving performance on a static corpus.

Like other learning programs, RIPPER takes as input the names of a set of *classes* to be learned, the names and ranges of values of a fixed set of *features*, and *training data* specifying the class and feature values for each example in a training set. Its output is a *classification model* for predicting the class of future examples. In RIPPER, the classification model is learned using greedy search guided by an information gain metric, and is expressed as an ordered set of if-then rules. By default RIPPER corrects for noisy data. In the experiments reported here, unlike those reported by Jordan and Walker (2000), corrections for noisy data have been suppressed since the reliability of the annotated features is high.

Thus to apply RIPPER, the object descriptions in the corpus are encoded in terms of a set of classes (the output classification), and a set of input features that are used as predictors for the classes. As mentioned above, the goal is to learn which of a set of content attributes should be included in an object description. Below we describe how a class is assigned to each object description, summarize the features extracted from the dialogue in which each expression occurs, and the method applied to learn to predict the class of object description from the features.

| Class Name | N in Corpus | Explicit attributes in object description |
|---|---|---|
| CPQ | 64 | Color, Price, Quantity |
| CPO | 56 | Color, Price, Owner |
| CPOQ | 46 | Color, Price, Owner, Quantity |
| T | 42 | None (type only) |
| CP | 41 | Color, Price |
| O | 32 | Owner |
| CO | 31 | Color, Owner |
| C | 18 | Color |
| CQ | 14 | Color, Quantity |
| COQ | 13 | Color, Owner, Quantity |
| OQ | 12 | Owner, Quantity |
| PO | 11 | Price, Owner |
| Q | 5 | Quantity |
| P | 4 | Price |
| PQ | 2 | Price, Quantity |
| POQ | 2 | Price, Owner, Quantity |

Figure 12: Encoding of attributes to be included in terms of ML Classes, ordered by frequency

## 4.1 Class Assignment

The corpus of object descriptions is used to construct the machine learning classes as follows. The learning task is to determine the subset of the four attributes, color, price, owner, quantity, to include in an object description. Thus one method for representing the class that each object description belongs to is to encode each object description as a member of the category represented by the set of attributes expressed by the object description. This results in 16 classes representing the power set of the four attributes as shown in Figure 12. The frequency of each class is also shown in Figure 12. Note that these classes are encodings of the hand annotated explicit attributes that were shown in Figure 6 but exclude the type attribute since we are not attempting to model pronominal selections.

## 4.2 Feature Extraction

The corpus is used to construct the machine learning features as follows. In RIPPER, feature values are continuous (numeric), set-valued, or symbolic. We encoded each discourse entity for a furniture item in terms of the set of 82 total features described in Section 3 as inspired by theories of content selection for subsequent reference. These features were either directly annotated by humans as described in Section 2, derived from annotated features, or inherent to the dialogue (Di Eugenio et al., 2000; Jordan, 2000b). The dialogue context in which each description occurs is directly represented in the encodings. In a dialogue system, the dialogue manager would have access to all these features, which are needed by the problem solving component, and would provide them to the language generator. The entire feature set is summarized in Figure 13.

- **Assumed Familiarity Features**

  – mutually known attributes: type-mk, color-mk, owner-mk, price-mk, quantity-mk

  – reference-relation: one of `initial`, `coref`, `set`, `class`, `cnanaphora`, `predicative`

- **Inherent Features**

  – utterance-number, speaker-pair, speaker, problem-number

  – attribute values:

    * type: one of `sofa`, `chair`, `table`, `rug`, `lamp`, `superordinate`
    * color: one of `red`, `blue`, `green`, `yellow`
    * owner: one of `self`, `other`, `ours`
    * price: range from $50 to $600
    * quantity: range from 0 to 4.

- **Conceptual Pact Features**

  – interactions with other discourse entities: distance-last-ref, distance-last-ref-in-turns, number-prev-mentions, speaker-of-last-ref, distance-last-related

  – previous description: color-in-last-exp, type-in-last-exp, owner-in-last-exp, price-in-last-exp, quantity-in-last-exp, type-in-last-turn, color-in-last-turn, owner-in-last-turn, price-in-last-turn, quantity-in-last-turn, initial-in-last-turn

  – frequency of attributes: freq-type-expressed, freq-color-expressed, freq-price-expressed, freq-owner-expressed, freq-quantity-expressed

  – stability history: cp-given-last-2, cp-given-last-3

- **Contrast Set Features**

  – distractor frequencies: type-distractors, color-distractors, owner-distractors, price-distractors, quantity-distractors

  – Attribute Saliency: majority-type, majority-type-freq, majority-color, majority-color-freq, majority-price, majority-price-freq, majority-owner, majority-owner-freq, majority-quantity, majority-quantity-freq

- **Intentional Influences Features**

  – task situation: goal, colormatch, colormatch-constraintpresence, pricelimit, pricelimit-constraintpresence, priceevaluator, priceevaluator-constraintpresence, colorlimit, colorlimit-constraintpresence, priceupperlimit, priceupperlimit-constraintpresence

  – agreement state: influence-on-listener, commit-speaker, solution-size, prev-influence-on-listener, prev-commit-speaker, prev-solution-size, distance-of-last-state-in-utterances, distance-of-last-state-in-turns, ref-made-in-prev-action-state, speaker-of-last-state, prev-ref-state

  – previous agreement state description: prev-state-type-expressed, prev-state-color-expressed, prev-state-owner-expressed, prev-state-price-expressed, prev-state-quantity-expressed

  – solution interactions: color-contrast, price-contrast

Figure 13: Full Feature Set for Representing Basis for Object Description Content Selection in Task Oriented Dialogues.

## 4.3 Learning Experiments

The final input for learning is training data, i.e., a representation of a set of discourse entities, their discourse context and their object descriptions in terms of feature and class values. In order to induce rules from a variety of feature representations, the training data is represented differently in different experiments.

The goal of these experiments is to test the contribution of the features suggested by the three models of object description content selection described in Section 3. Our prediction is that the INCREMENTAL and the INTENTIONAL INFLUENCES models will work best in combination for predicting object descriptions for both initial and subsequent reference. This is because: (1) the INTENTIONAL INFLUENCES features capture nothing relevant to the reference identification goal, which is the focus of the INCREMENTAL model, and (2) we hypothesize that the problem solving state will be relevant for selecting attributes for initial descriptions, and the INCREMENTAL model features capture nothing directly about the problem solving state, but this is the focus of the INTENTIONAL INFLUENCES model. Finally we expect the CONCEPTUAL PACT model to work best in conjunction with the INCREMENTAL and the INTENTIONAL INFLUENCES models since it is overriding informativeness constraints, and since, after establishing a pact, it may need to adapt the description to make it more efficient or re-negotiate the pact as the problem-solving situation changes.

Therefore, examples are first represented using only the ASSUMED FAMILIARITY features in Figure 7 to establish a performance baseline for assumed familiarity information. We then add individual feature sets to the ASSUMED FAMILIARITY feature set to examine the contribution of each feature set on its own. Thus, examples are represented using only the features specific to a particular model, i.e. CONCEPTUAL PACT features in Figure 11, the CONTRAST SET features in Figure 9 or the INTENTIONAL INFLUENCES features in Figure 10. Remember that there are three different versions of the CONTRAST SET features, derived from three different models of what is currently "in focus". One model (SEGMENT) is based on intentional structure (Grosz & Sidner, 1986). The other two are simple recency based models where the active focus space either contains only discourse entities from the most recent utterance or the most recent five utterances (ONE UTTERANCE, FIVE UTTERANCE).

In addition to the theoretically-inspired feature sets, we include the task and dialogue specific INHERENT features in Figure 8. These particular features are unlikely to produce rules that generalize to other domains, including new COCONUT dialogues, because each domain and pair of speakers will instantiate these values uniquely for a particular domain. Thus, these features may indicate aspects of individual differences, and the role of the specific situation in models of content selection for object descriptions.

Next, examples are represented using combinations of the features from the different models to examine interactions between feature sets.

Finally, to determine whether particular feature types have a large impact (e.g. frequency features), we report results from a set of experiments using singleton feature sets, where those features that varied by attribute alone are clustered into sets while the rest contain just one feature. For example, the distractor frequency attributes in Figure 9 form a cluster for a singleton feature set whereas *utterance-number* is the only member of its feature set. We experimented with singleton feature sets in order to determine if any are making a large impact on the performance of the model feature set to which they belong.

The output of each machine learning experiment is a model for object description generation for this domain and task, learned from the training data. To evaluate these models, the error rates of the learned models are estimated using 25-fold cross-validation, i.e. the total set of examples is randomly divided into 25 disjoint test sets, and 25 runs of the learning program are performed. Thus, each run uses the examples not in the test set for training and the remaining examples for testing. An estimated error rate is obtained by averaging the error rate on the test portion of the data from each of the 25 runs. For sample sizes in the hundreds (the COCONUT corpus provides 393 examples), cross-validation often provides a better performance estimate than holding out a single test set (Weiss & Kulikowski, 1991). The major advantage is that in cross-validation all examples are eventually used for testing, and almost all examples are used in any given training run.

## 5. Experimental Results

Table 2 summarizes the experimental results. For each feature set, and combination of feature sets, we report accuracy rates and standard errors resulting from 25-fold cross-validation. We test differences in the resulting accuracies using paired t-tests. The table is divided into regions grouping results using similar feature sets. Row 1 provides the accuracy for the MAJORITY CLASS BASELINE of 16.9%; this is the standard baseline that corresponds to the accuracy achieved from simply choosing the description type that occurs most frequently in the corpus, which in this case means that the object description generator would always use the color, price and quantity attributes to describe a domain entity. Row 2 provides a second baseline, namely that for using the ASSUMED FAMILIARITY feature set. This result shows that providing the learner with information about whether the values of the attributes for a discourse entity are mutually known does significantly improve performance over the MAJORITY CLASS BASELINE (t=2.4, p<.03). Examination of the rest of the table shows clearly that the accuracy of the learned object description generator depends on the features that the learner has available.

Rows 3 to 8 provide the accuracies of object description generators trained and tested using one of the additional feature sets in addition to the FAMILIARITY feature set. Overall, the results here show that compared to the FAMILIARITY baseline, the features for INTENTIONAL INFLUENCES (FAMILIARITY,INF t=10.0, p<.01), CONTRAST SET (FAMILIARITY,SEG t=6.1, p< .01; FAMILIARITY,1UTT t=4.7, p< .01; FAMILIARITY,5UTT t=4.2, p< .01), and CONCEPTUAL PACT (FAMILIARITY,CP t=6.2, p< .01) taken independently significantly improve performance. The accuracies for the INTENTIONAL INFLUENCES features (Row 7) are significantly better than for CONCEPTUAL PACT (t=5.2, p<.01) and the three parameterizations of the INCREMENTAL model (FAMILIARITY,SEG t=6, p<.01; FAMILIARITY,1UTT t=4.3, p<.01; FAMILIARITY,5UTT t=4.2, p<.01), perhaps indicating the importance of a direct representation of the problem solving state for this task.

In addition, interestingly, Rows 3, 4 and 5 show that features for the INCREMENTAL model that are based on the three different models of discourse structure all perform equally well, i.e. there are no statistically significant differences between the distractors predicted by the model of discourse structure based on intention (SEG) and the two recency based models (1UTT, 5UTT), even though the raw accuracies for distractors predicted by the intention-based model are typically higher.[6] The remainder of the table shows that the intention based model only performs better than a recency based model when it is combined with all features (Row 15 SEG vs. Row 16 1UTT t=2.1, p<.05).

---

6. This is consistent with the findings reported by Jordan (2000b) which used a smaller dataset to measure which discourse structure model best explained the data for the INCREMENTAL model.

| Row | Model Tested | Feature Sets Used | Accuracy (SE) |
|---|---|---|---|
| 1 | BASELINE | MAJORITY CLASS | 16.9% (2.1) |
| 2 | BASELINE | FAMILIARITY | 18.1% (2.1) |
| 3 | INCREMENTAL | FAMILIARITY,SEG | 29.0% (2.2) |
| 4 | INCREMENTAL | FAMILIARITY,1UTT | 29.0% (2.5) |
| 5 | INCREMENTAL | FAMILIARITY,5UTT | 30.4% (2.6) |
| 6 | CONCEPTUAL PACT | FAMILIARITY,CP | 28.9% (2.1) |
| 7 | INTENTIONAL INFLUENCES | FAMILIARITY,IINF | 42.4% (2.7) |
| 8 | SITUATION SPECIFIC | FAMILIARITY,INH | 54.5% (2.3) |
| 9 | INTENTIONAL INFLUENCES, INCREMENTAL | FAMILIARITY,IINF,SEG | 46.6% (2.2) |
| 10 | INTENTIONAL INFLUENCES, INCREMENTAL | FAMILIARITY,IINF,1UTT | 42.7% (2.2) |
| 11 | INTENTIONAL INFLUENCES, INCREMENTAL | FAMILIARITY,IINF,5UTT | 44.4% (2.6) |
| 12 | ALL THEORY FEATURES COMBINED | FAMILIARITY,IINF,CP,SEG | 43.2% (2.8) |
| 13 | ALL THEORY FEATURES COMBINED | FAMILIARITY,IINF,CP,1UTT | 40.9% (2.6) |
| 14 | ALL THEORY FEATURES COMBINED | FAMILIARITY,IINF,CP,5UTT | 41.9% (3.2) |
| 15 | ALL THEORIES & SITUATION SPECIFIC | FAMILIARITY,IINF,INH,CP,SEG | 59.9% (2.4) |
| 16 | ALL THEORIES & SITUATION SPECIFIC | FAMILIARITY,IINF,INH,CP,1UTT | 55.4% (2.2) |
| 17 | ALL THEORIES & SITUATION SPECIFIC | FAMILIARITY,IINF,INH,CP,5UTT | 57.6% (3.0) |
| 18 | BEST SINGLETONS FROM ALL MODELS COMBINED | FAMILIARITY,IINF,INH,CP,SEG | 52.9% (2.9) |
| 19 | BEST SINGLETONS FROM ALL MODELS COMBINED | FAMILIARITY,IINF,INH,CP,1UTT | 47.8% (2.4) |
| 20 | BEST SINGLETONS FROM ALL MODELS COMBINED | FAMILIARITY,IINF,INH,CP,5UTT | 50.3% (2.8) |

Table 2: Accuracy rates for the content selection component of a object description generator using different feature sets, SE = Standard Error. CP = the CONCEPTUAL PACT features. IINF = the INTENTIONAL INFLUENCES features. INH = the INHERENT features. SEG = the CONTRAST-SET, SEGMENT FOCUS SPACE features. 1UTT = the CONTRAST SET, ONE UTTERANCE FOCUS SPACE features, 5UTT = the CONTRAST SET, FIVE UTTERANCE FOCUS SPACE features.

Finally, the SITUATION SPECIFIC model based on the INHERENT feature set (Row 8) which is domain, speaker and task specific performs significantly better than the FAMILIARITY baseline (t=16.6, p< .01). It is also significantly better than any of the models utilizing theoretically motivated features. It is significantly better than the INTENTIONAL INFLUENCES model (t=5, p<.01), and the CONCEPTUAL PACT model (t=9.9, p<.01), as well as the three parameterizations of the INCREMENTAL model (SEG t=10, p<.01; 1UTT t=10.4, p<.01; 5UTT t=8.8, p<.01).

**Say** POQ **if** priceupperlimit-constraintpresence = IMPLICIT ∧ reference-relation = class

**Say** COQ **if** goal = SELECTCHAIRS ∧ colormatch-constraintpresence = IMPLICIT ∧ prev-solution-size = DETERMINATE ∧ reference-relation = coref

**Say** COQ **if** goal = SELECTCHAIRS ∧ distance-of-last-state-in-utterances >= 3 ∧ speaker-of-last-state = SELF ∧ reference-relation = initial

**Say** COQ **if** goal = SELECTCHAIRS ∧ prev-ref-state = STATEMENT ∧ influence-on-listener = action-directive ∧ prev-solution-size = DETERMINATE

**Say** C **if** prev-commit-speaker = commit ∧ influence-on-listener = action-directive ∧ color-contrast = no ∧ speaker-of-last-state = SELF

**Say** C **if** color-contrast = yes ∧ goal = SELECTTABLE ∧ prev-influence-on-listener = action-directive ∧ influence-on-listener = na

**Say** C **if** solution-size = DETERMINATE ∧ prev-influence-on-listener = na ∧ prev-state-color-expressed = yes ∧ prev-state-price-expressed = na ∧ prev-solution-size = DETERMINATE

**Say** CO **if** colorlimit = yes

**Say** CO **if** price-mk = yes ∧ prev-solution-size = INDETERMINATE ∧ price-contrast = yes ∧ commit-speaker = na

**Say** CO **if** price-mk = yes ∧ prev-ref-state = PARTNER-DECIDABLE-OPTION ∧ distance-of-last-state-in-utterances <= 1 ∧ prev-state-type-expressed = yes

**Say** O **if** prev-influence-on-listener = open-option ∧ reference-relation = coref

**Say** O **if** influence-on-listener = info-request ∧ distance-of-last-state-in-turns <= 0

**Say** CP **if** solution-size = INDETERMINATE ∧ price-contrast = yes ∧ distance-of-last-state-in-turns >= 2

**Say** CP **if** distance-of-last-state-in-utterances <= 1 ∧ goal = SELECTSOFA ∧ influence-on-listener = na ∧ reference-relation = class

**Say** T **if** prev-solution-size = DETERMINATE ∧ distance-of-last-state-in-turns <= 0 ∧ prev-state-type-expressed = yes ∧ ref-made-in-prev-action-state = yes

**Say** T **if** prev-solution-size = DETERMINATE ∧ colormatch-constraintpresence = EXPLICIT

**Say** T **if** prev-solution-size = DETERMINATE ∧ goal = SELECTSOFA ∧ prev-state-owner-expressed = na ∧ color-contrast = no

**Say** CPOQ **if** goal = SELECTCHAIRS ∧ prev-solution-size = INDETERMINATE ∧ price-contrast = no ∧ type-mk = no

**Say** CPOQ **if** distance-of-last-state-in-utterances >= 5 ∧ type-mk = no

**Say** CPOQ **if** goal = SELECTCHAIRS ∧ influence-on-listener = action-directive ∧ distance-of-last-state-in-utterances >= 2

**Say** CPO **if** influence-on-listener = action-directive ∧ distance-of-last-state-in-utterances >= 2 ∧ commit-speaker = offer

**Say** CPO **if** goal = SELECTSOFA ∧ distance-of-last-state-in-utterances >= 1

default **Say** CPQ

Figure 14: A Sampling of Rules Learned Using ASSUMED FAMILIARITY and INTENTIONAL INFLUENCES Features. The classes encode the four attributes, e.g CPOQ = Color,Price,Owner and Quantity, T = Type only

In Section 4.3, we hypothesized that the INCREMENTAL and INTENTIONAL INFLUENCES models would work best in combination. Rows 9, 10 and 11 show the results of this combination for each underlying model of discourse structure. Each of these combinations provides some increase in accuracy, however the improvements in accuracy over the object description generator based on the INTENTIONAL INFLUENCES features alone (Row 7) are not statistically significant.

Figure 14 shows the rules that the object description generator learns given the AS-SUMED FAMILIARITY and INTENTIONAL INFLUENCES features. The rules make use of both types of ASSUMED FAMILIARITY features and all four types of INTENTIONAL INFLUENCES features. The features representing mutually known attributes and those representing the attributes expressed in a previous agreement state can be thought of as overlapping with the CONCEPTUAL PACT model, while features representing problem-solving structure and agreement state may overlap with the INCREMENTAL model by indicating what is in focus.

One of the rules from the rule set in Figure 14 is:

> **Say** T **if** prev-solution-size = DETERMINATE ∧ colormatch-constraintpresence = EXPLICIT .

An example of a dialogue excerpt that matches this rule is shown in Figure 15. The rule captures a particular style of problem solving in the dialogue in which the conversants talk explicitly about the points involved in matching colors (*we only get 650 points without rug and bluematch in living room*) to argue for including a particular item (*rug*). In this case, because a solution had been proposed, the feature *prev-solution-size* has the value `determinate`. So the rule describes those contexts in which a solution has been counter-proposed, and support for the counter-proposal is to be presented.

---

D: I suggest that we buy my blue sofa 300, your 1 table high green 200, your 2 chairs red 50, my 2 chairs red 50 and you can decide the rest. What do you think

J: your 3 chair green my high table green 200 and my 1 chair green 100. your sofa blue 300 rug blue 250. we get 700 point. 200 for sofa in livingroom plus rug 10. 20 points for match. 50 points for match in dining room plus 20 for spending all. red chairs plus red table costs 600. we only get 650 points without **rug** and bluematch in living room. add it up and tell me what you think.

Figure 15: Example of a discourse excerpt that matches a rule in the INTENTIONAL INFLU-ENCES and ASSUMED FAMILIARITY rule set

---

Rows 12, 13 and 14 in Table 2 contain the results of combining the CONCEPTUAL PACT features with the INTENTIONAL INFLUENCES features and the CONTRAST SET features. These results can be directly compared with those in Rows 9, 10 and 11. Because *ripper* uses a heuristic search, the additional features have the effect of making the accuracies for the resulting models lower. However, none of these differences are statistically significant. Taken together, the results in Rows 9-14 indicate that the best accuracies obtainable without using situation specific features (the INHERENT feature set), is the combination of the INTENTIONAL INFLUENCES and CONTRAST SET features, with a best overall accuracy of 46.6% as shown in Row 9.

Rows 15, 16 and 17 contain the results for combining all the features, including the INHERENT feature set, for each underlying model of discourse structure. This time there is one significant difference between the underlying discourse models in which the intention-based model, SEGMENT, is significantly better than the ONE UTTERANCE recency model (t=2.1, p<.05) but not the FIVE UTTERANCE recency model. Of the models in this group only the SEGMENT model is significantly better than the models that use a subset of the features (vs. INHERENT t=2.4, p<.03). Figure 16 shows the generation rules learned with the best performing feature set shown in Row 15. Many task, entity and speaker specific features from the INHERENT feature set are used in these rules. This rule set performs at 59.9% accuracy, as opposed to 46.6% accuracy for the more general feature set (shown in Row 9). In this final rule set, no CONCEPTUAL PACT features are used and removing

**Say** Q **if** type=CHAIR ∧ price>=200 ∧ reference-relation=set ∧ quantity>=2.
**Say** Q **if** speaker=GARRETT ∧ color-distractors<=0 ∧ type=CHAIR.
**Say** PO **if** color=unk ∧ speaker-pair=GARRETT-STEVE ∧ reference-relation=initial ∧ color-contrast=no.
**Say** PO **if** majority-color-freq>=6 ∧ reference-relation=set.
**Say** PO **if** utterance-number>=39 ∧ type-distractors<=0 ∧ owner=SELF ∧ price>=100.
**Say** OQ **if** color=unk ∧ quantity>=2 ∧ majority-price-freq<=5.
**Say** OQ **if** prev-state-quantity-expressed=yes ∧ speaker=JULIE ∧ color=RED.
**Say** COQ **if** goal=SELECTCHAIRS ∧ price-distractors<=3 ∧ owner=SELF ∧ distance-of-last-state-in-utterances>=3 ∧ majority-price<=200.
**Say** COQ **if** quantity>=2 ∧ price<=-1 ∧ ref-made-in-prev-action-state=no.
**Say** COQ **if** quantity>=2 ∧ price-distractors<=3 ∧ quantity-distractors>=4 ∧ influence-on-listener=action-directive.
**Say** CQ **if** speaker-pair=DAVE-GREG ∧ utterance-number>=22 ∧ utterance-number<=27 ∧ problem<=1.
**Say** CQ **if** problem>=2 ∧ quantity>=2 ∧ price<=-1.
**Say** CQ **if** color=YELLOW ∧ quantity>=3 ∧ influence-on-listener=action-directive ∧ type=CHAIR.
**Say** C **if** price-mk=yes ∧ majority-type=SUPERORDINATE ∧ quantity-distractors>=3.
**Say** C **if** price-mk=yes ∧ utterance-number<=21 ∧ utterance-number>=18 ∧ prev-state-price-expressed=na ∧ majority-price>=200 ∧ color-distractors>=2.
**Say** CO **if** utterance-number>=16 ∧ price<=-1 ∧ type=CHAIR.
**Say** CO **if** price-mk=yes ∧ speaker-pair=JILL-PENNY.
**Say** CO **if** majority-price<=75 ∧ distance-of-last-state-in-utterances>=4 ∧ prev-state-type-expressed=na.
**Say** O **if** color=unk ∧ speaker-pair=GARRETT-STEVE.
**Say** O **if** color=unk ∧ owner=OTHER ∧ price<=300.
**Say** O **if** prev-influence-on-listener=open-option ∧ utterance-number>=22.
**Say** CP **if** problem>=2 ∧ quantity<=1 ∧ type=CHAIR.
**Say** CP **if** price>=325 ∧ reference-relation=class ∧ distance-of-last-state-in-utterances<=0.
**Say** CP **if** speaker-pair=JON-JULIE ∧ type-distractors<=1.
**Say** CP **if** reference-relation=set ∧ owner=OTHER ∧ owner-distractors<=0.
**Say** T **if** prev-solution-size=DETERMINATE ∧ price>=250 ∧ color-distractors<=5 ∧ owner-distractors>=2 ∧ utterance-number>=15.
**Say** T **if** color=unk.
**Say** T **if** prev-state-type-expressed=yes ∧ distance-of-last-state-in-turns<=0 ∧ owner-distractors<=4.
**Say** CPOQ **if** goal=SELECTCHAIRS ∧ prev-solution-size=INDETERMINATE.
**Say** CPOQ **if** speaker-pair=KATHY-MARK ∧ prev-solution-size=INDETERMINATE ∧ owner-distractors<=5.
**Say** CPOQ **if** goal=SELECTCHAIRS ∧ influence-on-listener=action-directive ∧ utterance-number<=22.
**Say** CPO **if** utterance-number>=11 ∧ quantity<=1 ∧ owner-distractors>=1.
**Say** CPO **if** influence-on-listener=action-directive ∧ price>=150.
**Say** CPO **if** reference-relation=class ∧ problem<=1.
default **Say** CPQ

Figure 16: A Sampling of the Best Performing Rule Set. Learned using the ASSUMED FAMILIARITY, INHERENT, INTENTIONAL INFLUENCES and CONTRAST SET feature sets. The classes encode the four attributes, e.g., CPOQ = Color,Price,Owner and Quantity, T = Type only.

these features during training had no effect on accuracy. All of the types of features in the ASSUMED FAMILIARITY, INHERENT, and CONTRAST SET are used. Of the INTENTIONAL INFLUENCES features, mainly the agreement state and previous agreement state descriptions are used. Some possible explanations are that the agreement state is a stronger influence than the task situation or that the task situation is not modelled well.

Why does the use of the INHERENT feature set contribute so much to overall accuracy and why are so many INHERENT features used in the rule set in Figure 16? It may be that the INHERENT features of objects would be important in any domain because there is a lot of domain specific reasoning in the task of object description content selection. However, these features are most likely to support rules that overfit to the current data set; as we have said before, rules based on the INHERENT feature set are unlikely to generalize to new situations. However, there might be more general or abstract versions of these features that could generalize to new situations. For example, the attribute values for the discourse entity may be capturing aspects of the problem solving (e.g. near the end of the problem, the price of expensive items is highly relevant). Second, the use of utterance-numbers can be characterized as rules about the beginning, middle and end of a dialogue and may again reflect problem solving progress. Third, the rules involving problem-number suggest that the behavior for the first problem is different from the others and may reflect that the dialogue partners have reached an agreement on their problem solving strategy. Finally, the use of speaker-pair features in the rules included all but two of the possible speaker-pairs, which may reflect differences in the agreements reached on how to collaborate. One of the rules from this rule set is shown below:

**Say** CP **if** price >= 325 ∧ reference-relation = class ∧ distance-of-last-state-in-utterances <= 0.

This rule applies to discourse entities in the dialogues of one speaker pair. An example dialogue excerpt that matches this rule is in Figure 17. The rule reflects a particular style of describing the items that are available to use in the problem solving, in which the speaker first describes the class of the items that are about to be listed. This style of description allows the speaker to efficiently list what he has available. The *distance-of-last-state-in-utterances* feature captures that this style of description occurs before any proposals have been made.

---

M: I have $550, my inventory consists of 2 Yellow hi-tables for $325 each. Sofas, **yellow for $400** and **green for $350**.

Figure 17: Example of a dialogue excerpt that matches a rule in the best performing rule set

---

As described above, we also created singleton feature sets, in addition to our theoretically inspired feature sets, to determine if any singleton features are, by themselves, making a large impact on the performance of the model it belongs to. The singleton features shown in Table 3 resulted in learned models that were significantly above the majority class baseline. The last column of Table 3 also shows that, except for the ASSUMED FAMILIARITY and INCREMENTAL 5UTT models, the theory model to which a particular singleton feature belongs is significantly better, indicating that no singleton alone is a better predictor than the combined features in these theoretical models. The ASSUMED FAMILIARITY and INCREMENTAL 5UTT models perform similarly to their corresponding single feature models indicating that these single features are the most highly useful features for these two models.

Finally, we combined all of the singleton features in Table 3 to learn three additional models shown in Rows 18, 19 and 20 of Table 2. These three models are not significantly different from one another. The best performing model in Row 15, which combines all features, is significantly better than 1UTT (t=4.2, p<.01) and 5UTT (t=2.8, p<.01) in Rows 19 and 20, but is not significantly different from SEG (t=2.0, NS) in Row 18. The combined singletons SEG model (Row 18) is also not significantly different from the INHERENT model (Row 8). The combined singletons SEG model has the advantage that it requires just two situation specific features and a smaller set of theoretical features.

| Source Model | Features in Set | Accuracy (SE) | Better than BASELINE | Source Model Better |
|---|---|---|---|---|
| ASSUMED FA- MILIARITY | type-mk, color-mk, owner-mk, price-mk, quantity-mk | 18.1% (2.1) | t=2.4, p<.03 | identical |
| CONCEPTUAL PACT | freq-type-expressed, freq-color-expressed, freq-price-expressed, freq-owner-expressed, freq-quantity-expressed | 22.1% (1.8) | t=3.7, p<.01 | t=5.7, p<.01 |
| | cp-given-last-2 | 20.9% (2.1) | t=3.9, p<.01 | t=4.3, p<.01 |
| | type-in-last-exp, color-in-last-exp, price-in-last-exp, owner-in-last-exp, quantity-in-last-exp | 18.9% (1.9) | t=2.8, p<.02 | t=5.7, p<.01 |
| | type-in-last-turn, color-in-last-turn, price-in-last-turn, owner-in-last-turn, quantity-in-last-turn | 18.1% (2.0) | t=3.4, p<.02 | t=6.4, p<.01 |
| INCREMENTAL SEG | type-distractors, color-distractors, price-distractors, owner-distractors, quantity-distractors | 21.4% (2.5) | t=3.2, p<.01 | t=3.6, p<.01 |
| | majority-type, majority-color, majority-price, majority-owner, majority-quantity | 19.9% (2.3) | t=2.5, p<.02 | t=4.8, p<.01 |
| INCREMENTAL 1UTT | type-distractors, color-distractors, price-distractors, owner-distractors, quantity-distractors | 20.8% (2.4) | t=3.2, p<.01 | t=3.2, p<.01 |
| INCREMENTAL 5UTT | type-distractors, color-distractors, price-distractors, owner-distractors, quantity-distractors | 25.7% (2.7) | t=4.4, p<.01 | t=1.5, NS |
| INTENTIONAL INFLUENCES | distance-of-last-state-in-utterances | 21.3% (2.0) | t=3.7, p<.01 | t=11, p<.01 |
| | distance-of-last-state-in-turns | 20.0% (2.1) | t=3.6, p<.01 | t=10.2, p<.01 |
| | colormatch | 19.3% (2.2) | t=3.7, p<.01 | t=10.3, p<.01 |
| | prev-state-type-expressed, prev-state-color-expressed, prev-state-owner-expressed, prev-state-price-expressed, prev-state-quantity-expressed | 19.2% (1.9) | t=3.6, p<.01 | t=8.8, p<.01 |
| SITUATION SPECIFIC | type, color, price, owner, quantity | 24.3% (2.5) | t=4.1, p<.01 | t=12.5, p<.01 |
| | utterance-number | 20.5% (2.3) | t=3.3, p<.01 | t=16.2, p<.01 |

Table 3: Performance using singleton feature sets, SE = Standard Error

| Class | recall | precision | fallout | F (1.00) |
|---|---|---|---|---|
| CPQ | 100.00 | 63.64 | 12.12 | 0.78 |
| CPO | 66.67 | 100.00 | 0.00 | 0.80 |
| CPOQ | 100.00 | 100.00 | 0.00 | 1.00 |
| T | 50.00 | 100.00 | 0.00 | 0.67 |
| CP | 100.00 | 100.00 | 0.00 | 1.00 |
| O | 100.00 | 60.00 | 5.41 | 0.75 |
| CO | 66.67 | 100.00 | 0.00 | 0.80 |
| C | 0.00 | 0.00 | 5.13 | 0.00 |
| CQ | 0.00 | 100.00 | 0.00 | 0.00 |
| COQ | 100.00 | 100.00 | 0.00 | 1.00 |
| PO | 50.00 | 100.00 | 0.00 | 0.67 |
| OQ | 66.67 | 50.00 | 5.41 | 0.57 |
| Q | 0.00 | 0.00 | 2.50 | 0.00 |
| POQ | 0.00 | 100.00 | 0.00 | 0.00 |
| PQ | 0.00 | 100.00 | 0.00 | 0.00 |

Table 4: Recall and Precision values for each class; the rows are ordered from most frequent to least frequent class

| Class | CPQ | O | COQ | C | CPO | CO | PO | T | OQ | POQ | CPOQ | Q | CP | PQ | CQ |
|---|---|---|---|---|---|---|---|---|---|---|---|---|---|---|---|
| CPQ | 7 | 0 | 0 | 0 | 0 | 0 | 0 | 0 | 0 | 0 | 0 | 0 | 0 | 0 | 0 |
| O | 0 | 3 | 0 | 0 | 0 | 0 | 0 | 0 | 0 | 0 | 0 | 0 | 0 | 0 | 0 |
| COQ | 0 | 0 | 2 | 0 | 0 | 0 | 0 | 0 | 0 | 0 | 0 | 0 | 0 | 0 | 0 |
| C | 1 | 0 | 0 | 0 | 0 | 0 | 0 | 0 | 0 | 0 | 0 | 0 | 0 | 0 | 0 |
| CPO | 0 | 1 | 0 | 1 | 4 | 0 | 0 | 0 | 0 | 0 | 0 | 0 | 0 | 0 | 0 |
| CO | 1 | 0 | 0 | 0 | 0 | 2 | 0 | 0 | 0 | 0 | 0 | 0 | 0 | 0 | 0 |
| PO | 0 | 1 | 0 | 0 | 0 | 0 | 1 | 0 | 0 | 0 | 0 | 0 | 0 | 0 | 0 |
| T | 0 | 0 | 0 | 1 | 0 | 0 | 0 | 1 | 0 | 0 | 0 | 0 | 0 | 0 | 0 |
| OQ | 0 | 0 | 0 | 0 | 0 | 0 | 0 | 0 | 2 | 0 | 0 | 1 | 0 | 0 | 0 |
| POQ | 0 | 0 | 0 | 0 | 0 | 0 | 0 | 0 | 1 | 0 | 0 | 0 | 0 | 0 | 0 |
| CPOQ | 0 | 0 | 0 | 0 | 0 | 0 | 0 | 0 | 0 | 0 | 6 | 0 | 0 | 0 | 0 |
| Q | 0 | 0 | 0 | 0 | 0 | 0 | 0 | 0 | 0 | 0 | 0 | 0 | 0 | 0 | 0 |
| CP | 0 | 0 | 0 | 0 | 0 | 0 | 0 | 0 | 0 | 0 | 0 | 0 | 1 | 0 | 0 |
| PQ | 1 | 0 | 0 | 0 | 0 | 0 | 0 | 0 | 1 | 0 | 0 | 0 | 0 | 0 | 0 |
| CQ | 1 | 0 | 0 | 0 | 0 | 0 | 0 | 0 | 0 | 0 | 0 | 0 | 0 | 0 | 0 |

Table 5: Confusion matrix for a held-out test set; The row label indicates the class, while the column indicates how the token was classified automatically.

Tables 4 and 5 show the recall and precision for each class and a sample confusion matrix for one run of the best performing model with a held-out test-set consisting of 40 examples. Table 4 shows that the overall tendency is for both recall and precision to be higher for classes that are more frequent, and lower for the less frequent classes as one would expect. Table 5 shows there aren't any significant sources of confusion as the errors are spread out across different classes.

# 6. Discussion and Future Work

This article reports experimental results for training a generator to learn which attributes of a discourse entity to include in an object description. To our knowledge, this is the first reported experiment of a trainable content selection component for object description generation in dialogue. A unique feature of this study is the use of theoretical work in cognitive science on how speakers select the content of an object description. The theories we used to inspire the development of features for the machine learner were based on Brennan and Clark's (1996) model, Dale and Reiter's (1995) model and Jordan's (2000b) model. Because Dale and Reiter's model relies on a model of discourse structure, we developed features to represent Grosz and Sidner's (1986) model of discourse structure, as well as features representing two simple recency based models of discourse structure. The object description generators are trained on the COCONUT corpus of task-oriented dialogues. The results show that:

- The best performing learned object description generator achieves a 60% match to human performance as opposed to a 17% majority class baseline;

- The ASSUMED FAMILIARITY feature set improves performance over the baseline;

- Features specific to the task, speaker and discourse entity (the inherent feature set) provide significant performance improvements;

- The CONCEPTUAL PACT feature set developed to approximate Brennan and Clark's model of object description generation significantly improves performance over both the baseline and ASSUMED FAMILIARITY;

- The CONTRAST SET features developed to approximate Dale and Reiter's model significantly improve performance over both the baseline and ASSUMED FAMILIARITY;

- The INTENTIONAL INFLUENCES features developed to approximate Jordan's model are the best performing theoretically-inspired feature set when taken alone, and the combination of the INTENTIONAL INFLUENCES features with the CONTRAST SET features is the best performing of the theoretically-based models. This combined model achieves an accuracy of 46.6% as an exact match to human performance and holds more promise of being general across domains and tasks than those that include the inherent features.

- Tests using singleton feature sets from each model showed that frequency features and the attributes last used have the most impact in the CONCEPTUAL PACT model, the distractor set features are the most important for the INCREMENTAL models, and features related to state have the biggest impact in the INTENTIONAL INFLUENCES model. But none of these singleton features perform as well as the feature combinations in the related model.

- A model consisting of a combination of the best singleton features from each of the other models was not significantly different from the best learned object description generator and achieved a 53% match to human performance with the advantage of fewer situation specific features.

Thus the choice to use theoretically inspired features is validated, in the sense that every set of cognitive features improves performance over the baseline.

In previous work, we presented results from a similar set of experiments, but the best model for object description generation only achieved an accuracy of 50% (Jordan & Walker, 2000). The accuracy improvements reported here are due to a number of new features that we derived from the corpus, as well as a modification of the machine learning algorithm to respect the fact that the training data for these experiments is not noisy.

It is hard to say how good our best-performing accuracy of 60% actually is as this is one of the first studies of this kind. There are several issues to consider. First, the object descriptions in the corpus may represent just **one** way to describe the entity at that point in the dialogue, so that using human performance as a standard against which to evaluate the learned object description generators provides an overly rigorous test (Oberlander, 1998; Reiter, 2002). Furthermore, we do not know whether humans would produce identical object descriptions given the same discourse situation. A previous study of anaphor generation in Chinese showed that rates of match for human speakers averaged 74% for that problem (Yeh & Mellish, 1997), and our results are comparable to that. Furthermore, the results show that including features specific to speaker and attribute values improves performance significantly. Our conclusion is that it may be important to quantify the best performance that a human could achieve at matching the object descriptions in the corpus, given the complete discourse context and the identity of the referent. In addition, the difficulty of this problem depends on the number of attributes available for describing an object in the domain; the object description generator has to correctly make four different decisions to achieve an exact match to human performance. Since the COCONUT corpus is publicly available, we hope that other researchers will improve on our results.

Another issue that must be considered is the extent to which these experiments can be taken as a test of the theories that inspired the feature sets. There are several reasons to be cautious in making such interpretations. First, the models were developed to explain subsequent reference and not initial reference. Second, the feature sets cannot be claimed in any way to be complete. It is possible that other features could be developed that provide a better representation of the theories. Finally, RIPPER is a propositional learner, and the models of object description generation may not be representable by a propositional theory. For example, models of object description generation rely on a representation of the discourse context in the form of some type of discourse model. The features utilized here represent the discourse context and capture aspects of the discourse history, but these representations are not as rich as those used by a rule-based implementation. However it is interesting to note that whatever limitations these models may have, the automatically trained models tested here perform better than the rule-based implementations of these theoretical models, reported by Jordan (2000b).

Another issue is the extent to which these findings might generalize across domains. While this is always an issue for empirical work, one potential limitation of this study is that Jordan's model was explicitly developed to capture features specific to negotiation dialogues such as those in the COCONUT corpus. Thus, it is possible that the features inspired by that theory are a better fit to this data. Just as CONCEPTUAL PACT features are less prominent for the COCONUT data and that data thus inspired a new model, we expect to find that other types of dialogue will inspire additional features and feature representations.

Finally, a unique contribution of this work is the experimental comparison of different representations of discourse structure for the task of object description generation. We tested three representations of discourse structure, one represented by features derived from Grosz and Sidner's model, and two recency based representations. One of the most surprising results of this work is the finding that features based on Grosz and Sidner's model do not improve performance over extremely simple models based on recency. This could be due to issues discussed by Walker (1996a), namely that human working memory and processing limitations play a much larger role in referring expression generation and interpretation than would be suggested by the operations of Grosz and Sidner's focus space model. However it could also be due to much more mundane reasons, namely that it is possible (again) that the feature sets are not adequate representations of the discourse structure model differences, or that the differences we found would be statistically significant if the corpus were much larger. However, again the results on the discourse structure model differences reported here confirm the findings reported by Jordan (2000b), i.e. it was also true that the focus space model did not perform better than the simple recency models in Jordan's rule-based implementations.

In future work, we plan to perform similar experiments on different corpora with different communications settings and problem types (e.g. planning, scheduling, designing) to determine whether our findings are specific to the genre of dialogues that we examine here, or whether the most general models can be applied directly to a new domain. Related to this question of generality, we have created a binary attribute inclusion model using domain independent feature sets but do not yet have a new annotated corpus upon which to test it.

## Acknowledgments

Thanks to William Cohen for helpful discussions on the use of RIPPER for this problem, and to the three anonymous reviewers who provided many helpful suggestions for improving the paper.